\documentclass{article}
\usepackage{spconf,amsmath,graphicx}

\usepackage{graphicx} 
\usepackage{bm} 
\newcommand{\tabincell}[2]{\begin{tabular}{@{}#1@{}}#2\end{tabular}} 
\usepackage{scrextend} 
\usepackage{enumitem}
\setenumerate{noitemsep}

\makeatletter
\def\hlinew#1{%
	\noalign{\ifnum0=`}\fi\hrule \@height #1 \futurelet
	\reserved@a\@xhline}
\makeatother
\title{Adversarial Deep Structured Nets for \\ Mass Segmentation from Mammograms}
\name{Wentao Zhu$^{\star}$ \qquad Xiang Xiang$^{\dagger}$ \qquad Trac D. Tran$^{\dagger}$ \qquad Gregory D. Hager$^{\dagger}$ \qquad Xiaohui Xie$^{\star}$}
 \address{$^{\star}$University of California, Irvine \qquad $^{\dagger}$Johns Hopkins University \\
 \{wentaoz1,xhx\}@ics.uci.edu, \{xxiang, trac,  hager\}@jhu.edu}
\begin{document}
%
\maketitle
\begin{abstract}
  Mass segmentation provides effective morphological features which are important for mass diagnosis. In this work, we propose a novel end-to-end network for mammographic mass segmentation which employs a fully convolutional network (FCN) to model a potential function, followed by a conditional random field (CRF) to perform structured learning. Because the mass distribution varies greatly with pixel position, the FCN is combined with a position priori. Further, we employ adversarial training to eliminate over-fitting due to the small sizes of mammogram datasets. Multi-scale FCN is employed to improve the segmentation performance. Experimental results on two public datasets, INbreast and DDSM-BCRP, demonstrate that our end-to-end network achieves better performance than state-of-the-art approaches. \footnote{https://github.com/wentaozhu/adversarial-deep-structural-networks.git}
\end{abstract}
\section{Introduction}
\label{sec:intro}
According to the American Cancer Society, breast cancer is the most frequently diagnosed solid cancer and the second leading cause of cancer death among U.S. women. Mammogram screening has been proven to be an effective way for early detection and diagnosis, which significantly decrease breast cancer mortality. Mass segmentation provides morphological features, which play crucial roles for diagnosis. 

Traditional studies on mass segmentation rely heavily on hand-crafted features. Model-based methods build classifiers and learn features from masses \cite{beller2005example,cardoso2015closed}. There are few works using deep networks for mammogram~\cite{deepmil}. Dhungel et al. employed multiple deep belief networks (DBNs), Gaussian mixture model (GMM) classifier and a priori as potential functions, and structured support vector machine (SVM) to perform segmentation \cite{dhungel2015deep}. They further used CRF with tree re-weighted belief propagation to boost the segmentation performance \cite{dhungel2015tree}. A recent work used the output from a convolutional network (CNN) as a complimentary potential function, yielding the state-of-the-art performance~\cite{dhungel2015deepmiccai}. However, the two-stage training used in these methods produces potential functions that easily over-fit the training data.

\begin{figure}[t]
	\begin{center}
		\begin{minipage}{\linewidth}
			\centerline{\includegraphics[width=6cm]{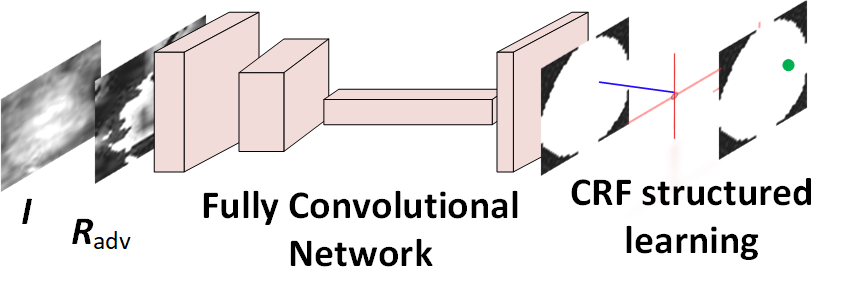}}
		\end{minipage}
		\caption{The proposed adversarial deep FCN-CRF network with four convolutional layers followed by CRF for structured learning.}
		\label{framework}
	\end{center}
\end{figure}

In this work, we propose an end-to-end trained adversarial deep structured network to perform mass segmentation (Fig. \ref{framework}). The proposed network is designed to robustly learn from a small dataset with poor contrast mammographic images. Specifically, an end-to-end trained FCN with CRF is applied. Adversarial training is introduced into the network to learn robustly from scarce mammographic images. Different from DI2IN-AN using a generative framework \cite{yang2017automatic}, we directly optimize pixel-wise labeling loss. To further explore statistical property of mass regions, a spatial priori is integrated into FCN. We validate the adversarial deep structured network on two public mammographic mass segmentation datasets. The proposed network is demonstrated to outperform other algorithms for mass segmentation consistently. 

Our main contributions in this work are: (1) We propose an unified end-to-end training framework integrating FCN+CRF and adversarial training. (2) We employ an end-to-end network to do mass segmentation while previous works require a lot of hand-designed features or multi-stage training. (3) Our model achieves the best results on two most commonly used mammographic mass segmentation datasets.
\section{FCN-CRF Network} 
Fully convolutional network (FCN) is a commonly used model for image segmentation, which consists of convolution, transpose convolution, or pooling \cite{long2015fully}. For training, the FCN optimizes maximum likelihood loss function 
\begin{equation}
\mathcal{L}_{FCN} = -\frac{1}{N\times N_i} \sum_{n=1}^{N}\sum_{i=1}^{N_i} \log P_{fcn}{(y_{n,i} | \textbf{I}_n; \bm \theta)},
\end{equation}
where $y_{n,i}$ is the label of $i$th pixel in the $n$th image $\textbf{I}_n$, $N$ is the number of training mammograms, $N_i$ is the number of pixels in the image, and $\bm \theta$ is the parameter of FCN. Here the size of images is fixed to $40 \times 40$ and $N_i$ is 1,600. 

CRF is a classical model for structured learning, well suited for image segmentation. It models pixel labels as random variables in a Markov random field conditioned on an observed input image. To make the annotation consistent, we use $\textbf{y} = (y_1, y_2, \dots, y_i, \dots, y_{1,600})^T$ to denote the random variables of pixel labels in an image, where $y_i\in\{0,1\}$. The zero denotes pixel belonging to background, and one denotes it belonging to mass region. The Gibbs energy of fully connected pairwise CRF is \cite{krahenbuhl2011efficient}
\begin{equation}
E(\textbf{y}) = \sum_{i} \psi_u (y_i) + \sum_{i<j} \psi_p (y_i, y_j),
\end{equation}
where unary potential function $\psi_u (y_i)$ is the loss of FCN in our case, pairwise potential function $\psi_p (y_i, y_j)$ defines the cost of labeling pair $(y_i, y_j)$,
\begin{equation}
\label{equ:pairwisepotential}
\psi_p (y_i, y_j) = \mu (y_i, y_j) \sum_{m} w^{(m)} k_G^{(m)}(\textbf{f}_i, \textbf{f}_j),
\end{equation}
where label compatibility function $\mu$ is given by the Potts model in our case, $w^{(m)}$ is the learned weight, pixel values $I_i$ and positions $p_i$ can be used as the feature vector $\textbf{f}_i$, $k_G^{(m)}$ is the Gaussian kernel applied to feature vectors \cite{krahenbuhl2011efficient},
\begin{equation}
\label{equ:gaussiankernel}
k_G(\textbf{f}_i, \textbf{f}_j) = [\exp(-|I_i - I_j|^2/2), \exp(-|p_i - p_j|^2)/2]^T.
\end{equation} 
Efficient inference algorithm can be obtained by mean field approximation $Q(\textbf{y}) = \prod_i Q_i(y_i) $ \cite{krahenbuhl2011efficient}. The update rule is
\begin{equation}
\label{equ:messagepassing}
\begin{aligned}
&\tilde{Q}_i^{(m)}(l)\leftarrow \sum_{i\neq j}k_G^{(m)}(\textbf{f}_i,\textbf{f}_j)Q_j(l) \text{ for all } m,\\
&\check{Q}_i(l)\leftarrow \sum_m w^{(m)}\tilde{Q}_i^{(m)}(l), \quad \hat{Q}_i(l)\leftarrow \sum_{l^\prime\in\mathcal{L}}\mu(l,l^\prime)\check{Q}_i(l),\\
&\breve{Q}_i(l)\leftarrow \exp(- \psi_u (y_i = l))-\hat{Q}_i(l), \quad Q_i\leftarrow \frac{1}{Z_i}\exp\left(\breve{Q}_i(l)\right),
\end{aligned}
\end{equation}
where the first equation is the message passing from label of pixel $i$ to label of pixel $j$, the second equation is re-weighting with the learned weights $w^{(m)}$, the third equation is compatibility transformation, the fourth equation is adding unary potentials, and the last step is normalization. Here $\mathcal{L} = \{0,1\}$ denotes background or mass. The initialization of inference employs unary potential function as $Q_i(y_i) = \frac{1}{Z_i} \exp(- \psi_u (y_i))$. The mean field approximation can be interpreted as a recurrent neural network (RNN) \cite{zheng2015conditional}. 
\section{Adversarial FCN-CRF Nets}
\begin{figure}[t]
	\begin{center}
		\begin{minipage}{0.3\linewidth}
			\centerline{\includegraphics[width=1cm]{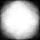}	\includegraphics[width=1cm]{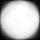}}
			\center{(a)}
		\end{minipage}
		\begin{minipage}{0.6\linewidth}
			\centerline{
				\includegraphics[width=1cm]{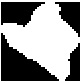}
				\includegraphics[width=1cm]{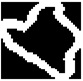}
				\includegraphics[width=1cm]{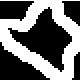}}
			\center{(b)}
		\end{minipage}
		\caption{The empirical estimation of a priori on INbreast (left) and DDSM-BCRP (right) training datasets (a). Trimap visualizations on the DDSM-BCRP dataset, segmentation groundtruth (first column), trimap of width $2$ (second column), trimaps of width $3$ (third column) (b).}
		\label{prior}
	\end{center}
\end{figure}

The shape and appearance priori play important roles in mammogram mass segmentation \cite{jiang2016mammographic,dhungel2015deepmiccai}. The distribution of labels varies greatly with position in the mammographic mass segmentation. From observation, most of the masses are located in the center of region of interest (ROI), and the boundary areas of ROI are more likely to be background (Fig. \ref{prior}(a)).

The conventional FCN provides independent pixel-wise predictions. It considers global class distribution difference corresponding to bias in the last layer. Here we employ a priori for position into consideration 
\begin{equation}
P(y_i | \textbf{I}; \bm \theta) \propto w_i P_{fcn}{(y_i | \textbf{I}; \bm \theta)},
\end{equation}
where $w_i$ is the empirical estimation of mass varied with the pixel position $i$, and $P_{fcn}{(y_i | \textbf{I}; \bm \theta)}$ is the predicted mass probability of conventional FCN. In the implementation, we added an image sized bias in the softmax layer as the empirical estimation of mass for FCN to train network. The $-\log P(y_i | \textbf{I}; \bm{\theta})$ is used as the unary potential function for $\psi_u (y_i)$ in the CRF as RNN. For multi-scale FCN as potential functions, the potential function is defined as $\psi_u (y_i) = \sum_{u ^{\prime}} w_{(u ^{\prime})} \psi_{u ^{\prime}}(y_i)$, where $w_{(u ^{\prime})}$ is the learned weight for unary potential function, $\psi_{u ^{\prime}}(y_i)$ is the  potential function provided by FCN of each scale. 

Adversarial training provides strong regularization for deep networks. The idea of adversarial training is that if the model is robust enough, it should be invariant to small perturbations of training examples that yield the largest increase in the loss (adversarial examples \cite{szegedy2013intriguing}). The perturbation $\bm{R}$ can be obtained as $\min_{\textbf{R}, \| \textbf{R} \| \leq \epsilon} \log P(\bm{y} | \textbf{I} + \textbf{R}; \bm \theta)$. In general, the calculation of exact $\bm R$ is intractable especially for complicated models such as deep networks. The linear approximation and $L_2$ norm box constraint can be used for the calculation of perturbation as $\bm{R}_{adv} = - \frac{\epsilon \bm{g}}{\|\bm{g}\|_2}$,
where $\bm{g} = \nabla_{\bm{I}} \log P(\bm{y} | \bm{I}; \bm{\theta})$. For adversarial FCN, the network predicts label of each pixel independently as $p(\bm{y}| \bm{I}; \bm{\theta}) = \prod_{i} P(y_i | \bm{I}; \bm{\theta})$. For adversarial CRF as RNN, the prediction of network relies on mean field approximation inference as $p(\bm{y}| \bm{I}; \bm{\theta}) = \prod_{i} Q(y_i | \bm{I}; \bm{\theta})$.

The adversarial training forces the model to fit examples with the worst perturbation direction. The adversarial loss is 
\begin{equation}
\label{equ:adversarialloss}
\mathcal{L}_{adv}(\bm{\theta}) = - \frac{1}{N} \sum_{n=1}^{N} \log P(\bm{y}_n | \bm{I}_n + \bm{R}_{adv,n}; \bm{\theta}).
\end{equation}
In training, the total loss is defined as the sum of adversarial loss and the empirical loss based on training samples as 
\begin{equation}
\label{equ:totalloss}
\mathcal{L}(\bm{\theta}) = \mathcal{L}_{adv}(\bm{\theta})- \frac{1}{N} \sum_{n=1}^{N} \log P(\bm{y}_n | \bm{I}_n; \bm{\theta}) + \frac{\lambda}{2} \| \bm{\theta} \|^2,
\end{equation}
where $\lambda$ is the \(l_2\) regularization factor for \(\bm \theta \), $P(\bm{y}_n | \bm{I}_n; \bm{\theta})$ is either mass probability prediction in the FCN or a posteriori approximated by mean field inference in the CRF as RNN for the $n$th image $\bm{I}_n$. 
\section{Experiments}\label{sec:exp}
We validate the proposed model on two most commonly used public mammographic mass segmentation datasets: INbreast \cite{moreira2012inbreast} and DDSM-BCRP dataset \cite{heath1998current}. We use the same ROI extraction and resize principle as \cite{dhungel2015deep,dhungel2015deepmiccai,dhungel2015tree}. Due to the low contrast of mammograms, image enhancement technique is used on the extracted ROI images as the first 9 steps in \cite{ball2007digital}, followed by pixel position dependent normalization. The preprocessing makes training converge quickly. We further augment each training set by flipping horizontally, flipping vertically, flipping horizontally and vertically, which makes the training set 4 times larger than the original training set. 

For consistent comparison, the Dice index metric is used to evaluate segmentation performance and is defined as $\frac{2 \times TP}{2 \times TP + FP + FN}$. For a fair comparison, we re-implement a two-stage model \cite{dhungel2015deepmiccai}, and obtain similar result (Dice index $0.9010$) on the INbreast dataset. 
\begin{itemize}[noitemsep]
	\item FCN is the network integrating a position priori into FCN (denoted as FCN 1 in Table \ref{tab:network}). 
	\item Adversarial FCN is FCN  with adversarial training.
	\item Joint FCN-CRF is the FCN followed by CRF as RNN with an end-to-end training scheme. 
	\item Adversarial FCN-CRF is the Jointly FCN-CRF with end-to-end adversarial training. 
	\item Multi-FCN, Adversarial multi-FCN, Joint multi-FCN-CRF, Adversarial multi-FCN-CRF employ 4 FCNs with multi-scale kernels, which can be trained in an end-to-end way using the last prediction. 
\end{itemize}
The prediction of Multi-FCN, Adversarial multi-FCN is the average prediction of the 4 FCNs. The configurations of FCNs are in Table~\ref{tab:network}. Each convolutional layer is followed by \(2\times2\) max pooling. The last layers of the four FCNs are all two $40\times40$ transpose convolution kernels with soft-max activation function. We use hyperbolic tangent activation function in middle layers. The parameters of FCNs are set such that the number of each layer's parameters is almost the same as that of CNN used in the work \cite{dhungel2015deepmiccai}. We use Adam with learning rate 0.003. The $\lambda$ is $0.5$ in the two datasets. The $\epsilon$ used in adversarial training are $0.1$ and $0.5$ for INbreast and DDSM-BCRP datasets respectively. Because the boundaries of masses on the DDSM-BCRP dataset are smoother than those on the INbreast dataset, we use larger perturbation $\epsilon$. For the CRF as RNN, we use 5 time steps in the training and 10 time steps in the test phase empirically. 
\begin{table}[t]
	\fontsize{9pt}{10pt}\selectfont\centering
	\caption{Kernel sizes of sub-nets (\#kernel$\times$\#width$\times$\#height).}\label{tab:network}
	\begin{tabular}{c|c|c|c}
		\hlinew{0.9pt}
		Net.&First layer&Second layer&Third layer\\
		\hline
		FCN 1&$6\times5\times5$&$12\times5\times5$ conv.&$588\times7\times7$\\
		\hline
		FCN 2&$9\times4\times4$&$12\times4\times4$ conv.&$588\times7\times7$\\
		\hline
		FCN 3&$16\times3\times3$& $13\times3\times3$ conv.&$415\times8\times8$\\
		\hline
		FCN 4&$37\times2\times2$&$12\times2\times2$ conv.&$355\times9\times9$\\
		\hlinew{0.9pt}
	\end{tabular}
\end{table}

\begin{table}[t]
	\fontsize{9pt}{10pt}\selectfont\centering
	\caption{Dices (\%) on INbreast and DDSM-BCRP datasets.}\label{tab:inbreast}
	\begin{tabular}{c|c|c}
		\hlinew{0.9pt}
		Methodology&INbreast&DDSM-BCRP\\		
		\hlinew{0.7pt}
		\tabincell{c}{Cardoso et al. \cite{cardoso2015closed}}&88&N/A \\
		\hline
		\tabincell{c}{Beller et al. \cite{beller2005example}}&N/A&70\\
		\hline
		\tabincell{c}{Deep Structure Learning \cite{dhungel2015deep}}&88&87\\
		\hline
		\tabincell{c}{TRW Deep Structure Learning \cite{dhungel2015tree}}&89&89\\
		\hline
		\tabincell{c}{Deep Structure Learning + CNN \cite{dhungel2015deepmiccai}}&90&90\\
		\hlinew{0.9pt}
		FCN & 89.48 & 90.21 \\
		\hline
		\tabincell{c}{ Adversarial FCN} & 89.71 & 90.78\\
		\hline
		\tabincell{c}{Joint FCN-CRF} & 89.78 & 90.97\\
		\hline
		\tabincell{c}{ Adversarial FCN-CRF}& 90.07 & 91.03\\
		\hline
		\tabincell{c}{Multi-FCN} & 90.47 & 91.17\\
		\hline
		\tabincell{c}{Adversarial multi-FCN} & 90.71 & 91.20\\
		\hline
		\tabincell{c}{Joint multi-FCN-CRF} & 90.76 & 91.26\\
		\hline
		\tabincell{c}{Adversarial multi-FCN-CRF} & \textbf{90.97} & \textbf{91.30}\\
		\hlinew{0.9pt}
	\end{tabular}
\end{table}
The INbreast dataset is a recently released mammographic mass analysis dataset, which provides more accurate contours of lesion region and the mammograms are of high quality. For mass segmentation, the dataset contains 116 mass regions. We use the first 58 masses for training and the rest for test, which is of the same protocol as \cite{dhungel2015deep,dhungel2015deepmiccai,dhungel2015tree}. The DDSM-BCRP dataset contains 39 cases (156 images) for training and 40 cases (160 images) for testing~\cite{heath1998current}. After ROI extraction, there are 84 ROIs for training, and 87 ROIs for test. We compare schemes with other recently published mammographic mass segmentation methods in Table \ref{tab:inbreast}. 

Table \ref{tab:inbreast} shows the CNN features provide superior performance on mass segmentation, outperforming hand-crafted feature based methods \cite{cardoso2015closed,beller2005example}. Our enhanced FCN achieves 0.25\% Dice index improvement than the traditional FCN on the INbreast dataset. The adversarial training yields 0.4\% improvement on average. Incorporating the spatially structured learning further produces 0.3\% improvement. Using multi-scale model contributes the most to segmentation results, which shows multi-scale features are effective for pixel-wise classification in mass segmentation. Combining all the components together achieves the best performance with 0.97\%, 1.3\% improvement on INbreast, DDSM-BCRP datasets respectively. The possible reason for the improvement is adversarial scheme eliminates the over-fitting.
We calculate the p-value of McNemar’s Chi-Square Test to compare our model with ~\cite{dhungel2015deepmiccai} on the INbreast dataset. We obtain p-value $< 0.001$, which shows our model is significantly better than
model~\cite{dhungel2015deepmiccai}. 

\begin{figure}[t]
	\begin{center}
		\begin{minipage}{\linewidth}
			\centerline{\includegraphics[width=8cm]{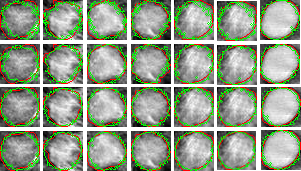}}
		\end{minipage}
		\caption{Visualization of segmentation results using the FCN (first row), Adversarial FCN (second row), Joint FCN-CRF (third row), Adversarial FCN-CRF (fourth row) on the test sets of INbreast dataset. Each column denotes a test sample. Red lines denote the ground truth. Green lines or points denote the segmentation results. Adversarial training provides sharper and more accurate segmentation boundaries than methods without adversarial training.}
		\label{fig:all}
	\end{center}
\end{figure}

To better understand the adversarial training, we visualize segmentation results in Fig. \ref{fig:all}. We observe that the segmentations in the second and fourth rows have more accurate boundaries than those of the first and third rows. It demonstrates the adversarial training improves FCN and FCN-CRF.

\begin{figure}[t]
	\begin{center}
		\begin{minipage}{0.46\linewidth}
			\centerline{\includegraphics[width=4.713cm]{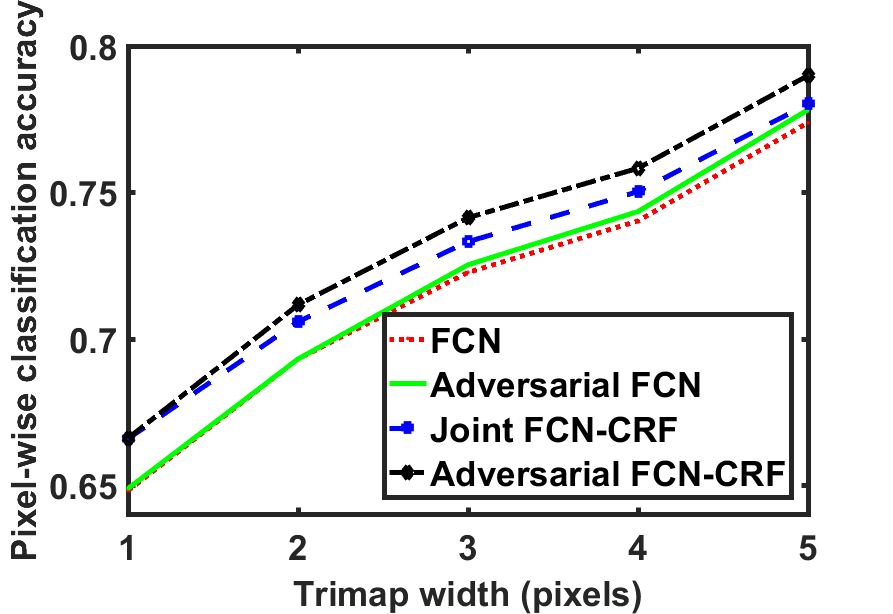}}
			\center{(a)}
		\end{minipage}
		\hspace{0.45cm}
		\begin{minipage}{0.46\linewidth}
			\centerline{\includegraphics[width=4.713cm]{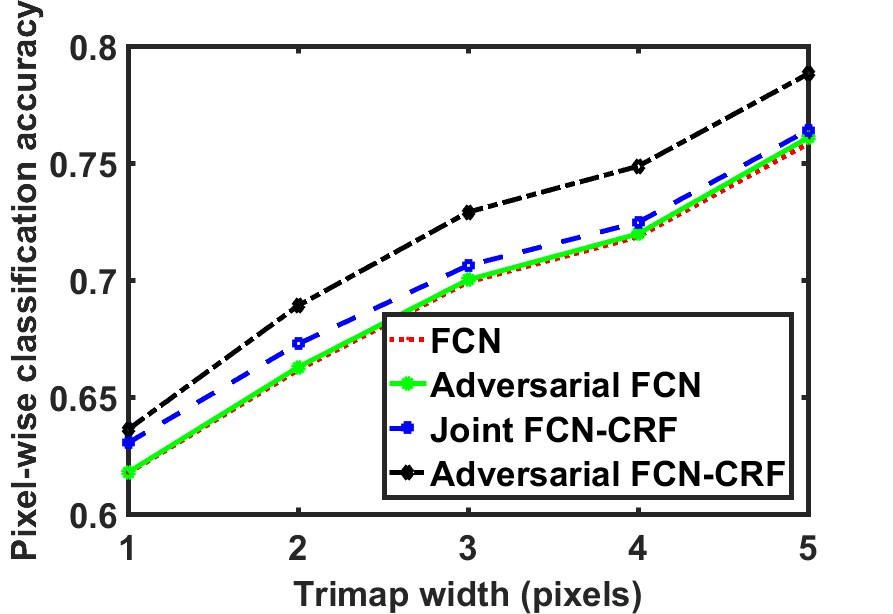}}
			\center{(b)}
		\end{minipage}
		\caption{Accuracy comparisons among FCN, Adversarial FCN, Joint FCN-CRF and Adversarial FCN-CRF in trimaps with pixel width $1$, $2$, $3$, $4$, $5$ on the INbreast dataset (a) and the DDSM-BCRP dataset (b). The adversarial training improves segmentation accuracy around boundaries.}
		\label{fig:trimapinbreast}
	\end{center}
\end{figure}

We further employ the prediction accuracy based on trimap to specifically evaluate segmentation accuracy in boundaries \cite{kohli2009robust}. We calculate the accuracies within trimap surrounding the actual mass boundaries (groundtruth) in Fig. \ref{fig:trimapinbreast}. Trimaps on the DDSM-BCRP dataset is visualized in Fig.~\ref{prior}(b). From the figure, accuracies of Adversarial FCN-CRF are 2-3 \% higher than those of Joint FCN-CRF on average and the accuracies of Adversarial FCN are better than those of FCN. The above results demonstrate that the adversarial training improves the FCN and Joint FCN-CRF both for whole image and boundary region segmentation. 
\section{Conclusion}\label{sec:con}
In this work, we propose an end-to-end adversarial FCN-CRF network for mammographic mass segmentation. To integrate the priori distribution of masses and fully explore the power of FCN, a position priori is added to the network. Furthermore, adversarial training is used to handle the small size of training data by reducing over-fitting and increasing robustness. Experimental results demonstrate the superior performance of adversarial FCN-CRF on two commonly used public datasets.

\bibliographystyle{IEEEbib}
\small \bibliography{aaai}

\begin{thebibliography}{10}

\bibitem{beller2005example}
M.~Beller et~al.,
\newblock ``An example-based system to support the segmentation of stellate
  lesions,''
\newblock Springer, 2005.

\bibitem{cardoso2015closed}
J.~S~Cardoso et~al.,
\newblock ``Closed shortest path in the original coordinates with an
  application to breast cancer,''
\newblock {\em IJPRAI}, 2015.

\bibitem{deepmil}
W.~Zhu et~al,
\newblock ``Deep multi-instance networks with sparse label assignment for whole
  mammogram classification,''
\newblock {\em MICCAI}, 2017.

\bibitem{dhungel2015deep}
N.~Dhungel et~al.,
\newblock ``Deep structured learning for mass segmentation from mammograms,''
\newblock in {\em ICIP}. IEEE, 2015.

\bibitem{dhungel2015tree}
N.~Dhungel et~al.,
\newblock ``Tree re-weighted belief propagation using deep learning potentials
  for mass segmentation from mammograms,''
\newblock in {\em ISBI}. IEEE, 2015.

\bibitem{dhungel2015deepmiccai}
N.~Dhungel et~al.,
\newblock ``Deep learning and structured prediction for the segmentation of
  mass in mammograms,''
\newblock in {\em MICCAI}, 2015.

\bibitem{yang2017automatic}
D.~Yang et~al.,
\newblock ``Automatic liver segmentation using an adversarial image-to-image
  network,''
\newblock in {\em MICCAI}. Springer, 2017.

\bibitem{long2015fully}
J.~Long, E.~Shelhamer, and T.~Darrell,
\newblock ``Fully convolutional networks for semantic segmentation,''
\newblock in {\em CVPR}, 2015.

\bibitem{krahenbuhl2011efficient}
P.~Kr{\"a}henb{\"u}hl and V.~Koltun,
\newblock ``Efficient inference in fully connected crfs with gaussian edge
  potentials,''
\newblock in {\em NIPS}, 2011.

\bibitem{zheng2015conditional}
S.~Zheng et~al.,
\newblock ``Conditional random fields as recurrent neural networks,''
\newblock in {\em ICCV}, 2015.

\bibitem{jiang2016mammographic}
M.~Jiang et~al.,
\newblock ``Mammographic mass segmentation with online learned shape and
  appearance priors,''
\newblock in {\em MICCAI}, 2016.

\bibitem{szegedy2013intriguing}
C.~Szegedy et~al.,
\newblock ``Intriguing properties of neural networks,''
\newblock {\em ICLR}, 2014.

\bibitem{moreira2012inbreast}
I.~C~Moreira et~al.,
\newblock ``Inbreast: toward a full-field digital mammographic database,''
\newblock {\em Academic radiology}, 2012.

\bibitem{heath1998current}
M.~Heath et~al.,
\newblock ``Current status of the digital database for screening mammography,''
\newblock in {\em Digital mammography}. 1998.

\bibitem{ball2007digital}
J.~Ball et~al.,
\newblock ``Digital mammographic computer aided diagnosis using adaptive level
  set segmentation,''
\newblock in {\em EMBI}, 2007.

\bibitem{kohli2009robust}
P.~Kohli et~al.,
\newblock ``Robust higher order potentials for enforcing label consistency,''
\newblock {\em IJCV}, 2009.

\end{thebibliography}

\end{document}